\documentclass[sn-basic,Numbered,pdflatex]{sn-jnl}
\def\x{\mathbf{x}}
\def\e{\mathbf{e}}
\def\E{\mathbf{E}}
\def\h{\mathbf{h}}
\def\BN{\text{BN}}
\def\MH{\text{MH}}
\def\FF{\text{FF}}
\def\LN{\text{LN}}
\def\tour{\pmb{\pi}}
\newcommand{\rev}[1]{{\textcolor{black}{#1}}}
\usepackage{graphicx}%
\usepackage{multirow}%
\usepackage{amsmath,amssymb,amsfonts}%
\usepackage{amsthm}%
\usepackage{mathrsfs}%
\usepackage[title]{appendix}%
\usepackage{xcolor}%
\usepackage{textcomp}%
\usepackage{manyfoot}%
\usepackage{booktabs}\let\cline\cmidrule
\usepackage{algorithm}%
\usepackage{algorithmicx}%
\usepackage{algpseudocode}%
\usepackage{listings}%

\usepackage[normalem]{ulem}
\usepackage{array}
 \usepackage{upgreek}
\usepackage{pdfpages}
\usepackage{hyperref}
\usepackage{stfloats}
\useunder{\uline}{\ul}{}

\theoremstyle{thmstyleone}%
\theoremstyle{thmstyletwo}%

\theoremstyle{thmstylethree}%

\begin{document}

\title{A Lightweight CNN-Transformer Model for Learning Traveling Salesman Problems}


\author[1]{\fnm{Minseop} \sur{Jung}}\email{minseob22@inu.ac.kr}
\author[1]{\fnm{Jaeseung} \sur{Lee}}\email{hunni10@inu.ac.kr}
\author*[1]{\fnm{Jibum} \sur{Kim}}\email{jibumkim@inu.ac.kr}

%\author*[1,2]{\fnm{First} \sur{Author}}\email{iauthor@gmail.com}



\affil[1]{
\orgdiv{Department of Computer Science and Engineering},
\orgname{Incheon National University},
\orgaddress{\city{Incheon}, \postcode{22012}, \country{South Korea}}}

%\affil[2]{\orgdiv{Department}, \orgname{Organization}, \orgaddress{\street{Street}, \city{City}, \postcode{10587}, \state{State}, \country{Country}}}

%\affil[3]{\orgdiv{Department}, \orgname{Organization}, \orgaddress{\street{Street}, \city{City}, \postcode{610101}, \state{State}, \country{Country}}}


\abstract{Several studies have attempted to solve traveling salesman problems (TSPs) using various deep learning techniques. Among them, Transformer-based models show state-of-the-art performance even for large-scale Traveling Salesman Problems (TSPs). However, they are based on fully-connected attention models and suffer from  large computational complexity and GPU memory usage. 
\rev{Our work is the first CNN–Transformer model based on a CNN embedding layer and partial self-attention for TSP.}  %revision
 Our CNN-Transformer model is able to better learn spatial features from input data using a CNN embedding layer compared with the standard Transformer-based models. It also removes considerable redundancy in fully-connected attention models using the proposed partial self-attention. 
\rev{Experimental results show that the proposed CNN embedding layer and partial self-attention are very effective in improving performance and computational complexity. The proposed model exhibits the best performance in real-world datasets and outperforms other existing state-of-the-art (SOTA) Transformer-based models in various aspects.} %revision
 Our code is publicly  available at \url{https://github.com/cm8908/CNN_Transformer3}.}




\maketitle

\section{Introduction}\label{sec:intro}

The Traveling Salesman Problem (TSP) is a classic NP-Hard problem in computer science and operation research that seeks to find the shortest possible route to visit every city exactly once and return to the starting city~\cite{tsp-intro}. Finding an optimal solution for the TSP is computationally expensive when the number of cities is large. Researchers have studied a variety of heuristics and approximation algorithms that can provide high-quality solutions to the problem in a reasonable amount of time. 

One of the most simple and popular heuristics is a nearest-neighbor heuristics. It starts  at a randomly chosen city and repeatedly selects the nearest unvisited city as the next city to visit while there are unvisited cities. Finally, it returns to the starting city to complete the tour. Another famous heuristics for TSP is the Christofides algorithm~\cite{christofides}. It finds an approximate solution using the minimum spanning tree of the graph representing the cities, which is a tree that connects all the cities with the minimum possible total edge weight. It is known to provide a solution that is guaranteed to be within a factor of 3/2 of the optimal solution. Another famous tool for solving TSP using a heuristics approach is to use Google-OR tools~\cite{ortools}. It performs local search and meta-heuristics to find the approximate solutions of a wide range of combinatorial optimization problems such as TSP and vehicle routing problems. However, heuristic approaches trade optimality for computational cost and are expressed in the form of rules~\cite{kool,bresson}. 

\rev{For many TSP instances, Concorde is considered as the fastest and most exact TSP solver that produces the optimal solution \cite{concorde}.} %revision
Concorde uses an Integer Programming solver with Cutting Planes and Branch-and-Bound. It assumes  a symmetric TSP where the distance between two cities is the same in each opposite direction \cite{ptr-net}. Gurobi also finds optimal TSP results but Concorde is faster than Gurobi because  it is specialized for TSPs \cite{kool,gurobi}.

Many studies have been conducted to find an approximate solution of TSP based on deep learning. Among these studies, the  pioneering work is the Pointer Network \cite{ptr-net}. It is a supervised learning-based approach that uses RNN-based encoders and decoders. In the experiment, a planar symmetric TSP is assumed, and they use a beam search decoding procedure to remove invalid tours such as visiting the same city twice or ignoring a destination. Bello et al.  updated the learning parameters of the LSTM-based model with a reinforcement learning-based approach that uses the tour length as a negative reward signal \cite{bello}. Nazari et al. added an embedding instead of using the RNN encoder of the Pointer Network to reduce the computational complexity without impacting performance \cite{nazari}. Joshi et al. proposed a method for predicting the edge probability matrix of the entire graph through a graph convolutional neural network model and a supervised learning-based approach \cite{joshi}. Stohy et al. proposed a hybrid pointer network model for TSP that demonstrated  good performance for large-scale TSP instances \cite{hybrid-ptrnet}. However, it suffers from a long inference time towing to a more complex model structure compared to the baseline graph pointer network \cite{graph-ptrnet}. Several efforts have been made to introduce a convolutional neural network (CNN) to the TSP. Researchers used 2D convolution for TSP but did not show good performance \cite{miki-cnn,ling-cnn}. Sultana et al. introduced  a new model that combines 1D-CNN with LSTM but is still an RNN-based model \cite{sultana}.

Recent attention-based transformer models have shown good performance in various research fields \cite{vaswani,transformer-xl,bert,vi-transformer}. Researchers successfully used a transformer-based model to find approximate solutions for TSP \cite{deudon,wu,kool,pomo,bresson,goh,memory-eff}. Deudon et al. proposed a novel approach for solving TSP using deep reinforcement learning. The city coordinates are utilized as inputs, and the model is trained using reinforcement learning to predict a distribution of a city sequence \cite{deudon}. Kool et al. proposed a transformer-based model consisting  of purely attention blocks and trained the model using REINFORCE for solving various routing problems such as TSP and vehicle routing problems \cite{kool}. Wu et al. proposed a transformer-based deep reinforcement learning framework that trains an improvement heuristic that iteratively improves an initial solution \cite{wu}. Researchers proposed an approach that applied multiple rollout and data augmentation methods to Kool's attention model \cite{pomo}. 

Recently, Bresson et al. proposed a TSP Transformer model \cite{bresson}. It is based on a standard Transformer encoder with multi-head attention and residual connection but uses batch normalization instead of using layer normalization. It uses an auto-regressive decoding approach and introduces a self-attention block in the decoder part. It constructs the query using all cities in the partial tour with a self-attention module \cite{bresson}. They showed a state-of-the-art (SOTA) performance for various TSP instances and reported performance with an optimal (optimality) gap of 0.0004\% for TSP50 and 0.39\% for TSP100. 
\rev{Although the TSP Transformer model shows the SOTA for many TSP instances, it has a complex model structure based on a fully-connected attention-based model. It also requires large GPU usage.} %revision
 Moreover, the training and inference time are very long \cite{star-transformer}. Recently, various studies have been conducted to reduce the computational complexity of standard transformer models \cite{longformer,linformer,informer}. 
\rev{For TSP, a recent study has been conducted to make the model lightweight while removing the learnable decoder \cite{goh}.} %revision
\rev{A similar study to lightweight TSP Transformer model is performed in \cite{memory-eff}.} %revision 
\rev{Yang et al. proposed a memory-efficient Transformer-based model for TSP called Tspformer.} %revision
\rev{Their model successfully reduces GPU/CPU memory usage compared with the standard Transformer-based models \cite{memory-eff}, but the solution quality is not as good as the SOTA model.}

\rev{Pan et al. proposed a constructive approach based on hierarchical reinforcement learning (H-TSP), which is specialized in solving large-scale TSP instances \cite{h-tsp}. It employs a hierarchical deep reinforcement learning approach with policies in two levels: upper- and lower-level policies. While H-TSP demonstrates excellent performance on large-scale TSP instances, it requires an additional warm-up stage for the lower-level model by pre-training and undergoes a rather complex training process. Therefore, it suffers from the drawback of demanding substantial computing resources and significant training time. Ren et al. tackled the dynamic TSP using a self-supervised reinforcement learning approach \cite{ren}. This paper proposed a new feature extraction mechanism combining self-attention and context attention mechanisms \cite{ren}. This approach has the advantage of not requiring a manually crafted reward function.} %revision

In this paper, we propose a novel CNN-Transformer model based on partial self-attention by performing attention only on recently visited nodes in the decoder. Linear embedding in the standard Transformer model does not consider local spatial information and has limitations in learning local compositionality. 
\rev{Therefore, we add a CNN embedding layer to the standard Transformer model to extract the local spatial features of the input data, as the CNN is effective in learning the spatial invariance of nodes in the Euclidean space.} %revision
Second, the standard Transformer model is based on fully-connected attention-based models \cite{star-transformer}. Therefore, it suffers from huge computational complexity and memory consumption. Furthermore, the Transformer model structure has a weakness at learning local compositionality owing to  its fully-connected topology. For TSP, we improve the attention mechanism by proposing partial self-attention that focuses only on recently visited nodes in the decoder. 
\rev{Our observations reveal a significant reduction in redundancy in the Transformer model’s fully-connected topology for TSP. This reduction improves the TSP solution’s quality by removing excessive attention connections.} %revision
The main contributions of our paper are summarized as follows:

\begin{itemize}
\item To the best of our knowledge, we propose the first CNN-Transformer-based model for learning TSP solutions. Our results show that the CNN embedding layer is very effective in learning local spatial features of various TSP instances.
\item The proposed model is based on partial self-attention that performs attention only on recently visited nodes in the decoder. Therefore, the proposed model is able to better learn local compositionality compared with the standard Transformer model that is based on the fully-connected topology.
\item By removing the redundant attention connections in the decoder, the proposed model significantly reduces the GPU memory usage and has much  lower inference time compared with the standard Transformer model.
\end{itemize}

%%%%%%%%%%%%%%%%%%%%%%%%%
%%%%% PROPOSED MODEL %%%%
%%%%%%%%%%%%%%%%%%%%%%%%%
\section{Proposed Model}\label{sec:model}
%%%% Fig. 1 Overall structure %%%%
\begin{figure*}
    \centering
    \resizebox{\textwidth}{!}{
    \includegraphics[trim={0 0.6cm 1.2cm 1cm}, clip]{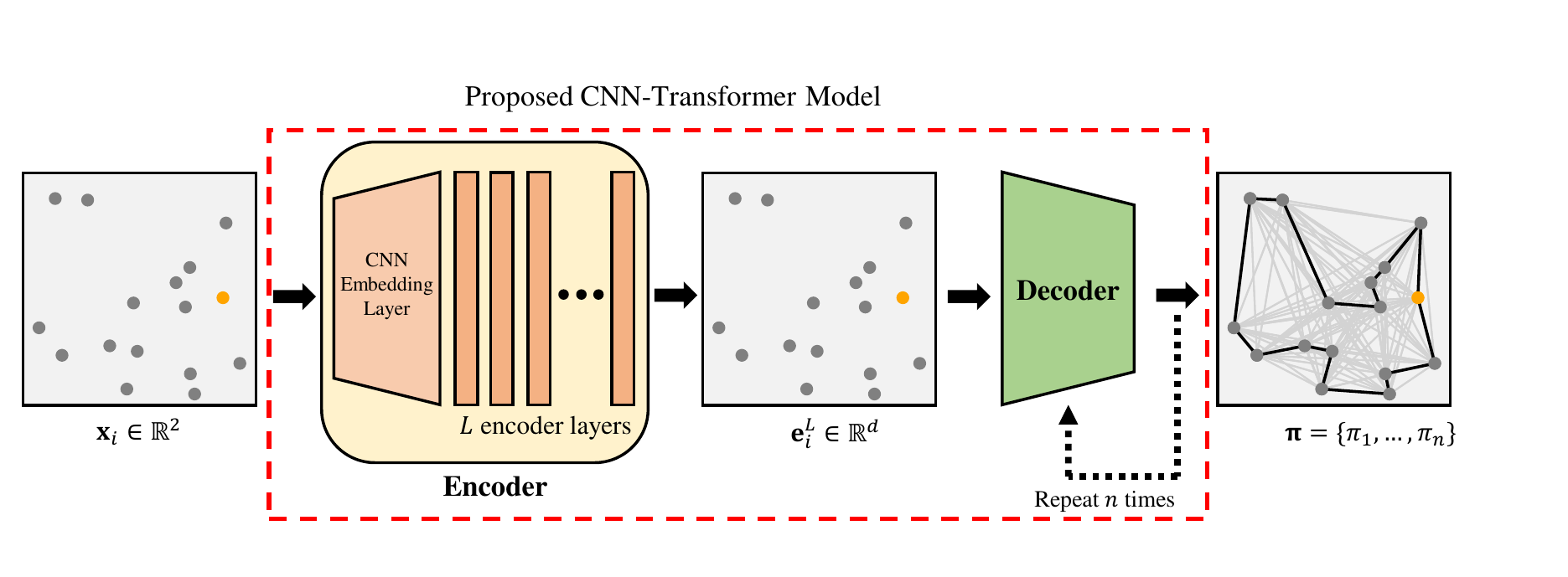}
    }
    \caption{Overview of the proposed CNN-Transformer model for TSP}
    \label{fig:model_overall}
\end{figure*}

We propose a CNN-Transformer model that combines a convolutional neural network  embedding layer with a standard Transformer model. 
\rev{The proposed CNN-Transformer model has the architecture of the encoder and decoder. It has a CNN embedding layer in the encoder to extract local spatial information.} %revision
 We improve the attention mechanism based on a partial self-attention, which removes unnecessary attention connections. 

Figure~\ref{fig:model_overall} illustrates the overall structure of the proposed model. The input of the proposed model is a planar point set $\mathbf{X}=\{\x_1,\dots ,\x_n\}$ with $n$ nodes (cities), where $\x_i\in\mathbb{R}^2$ represents the 2D Cartesian coordinates of the points. The output of the model, denoted as $\tour=\{\pi_1,\dots ,\pi_n\}$, is a sequence that represents the optimal predicted tour where $\pi_t$ is the node index selected at the $t^\text{th}$ decoding step at the decoder. Let $D(\x_i,\x_j)$ be the  distance between nodes $\x_i$ and $\x_j$. Our goal is to minimize the total tour length $\sum_{t=1}^{n-1} D(\x_{\pi_t},\x_{\pi_{t+1}})+D(\x_{\pi_n},\x_{\pi_1})$ while visiting each node exactly once and then return to the starting node.

%%%%%%%% ENCODER %%%%%%%%%%%
\subsection{Encoder}\label{subsec:encoder}

\begin{figure}[t!]
    \centering
    %\resizebox{\columnwidth}{!}
   {
    \includegraphics[trim={0 0.5cm 0 0.5cm}, clip, width=6cm]{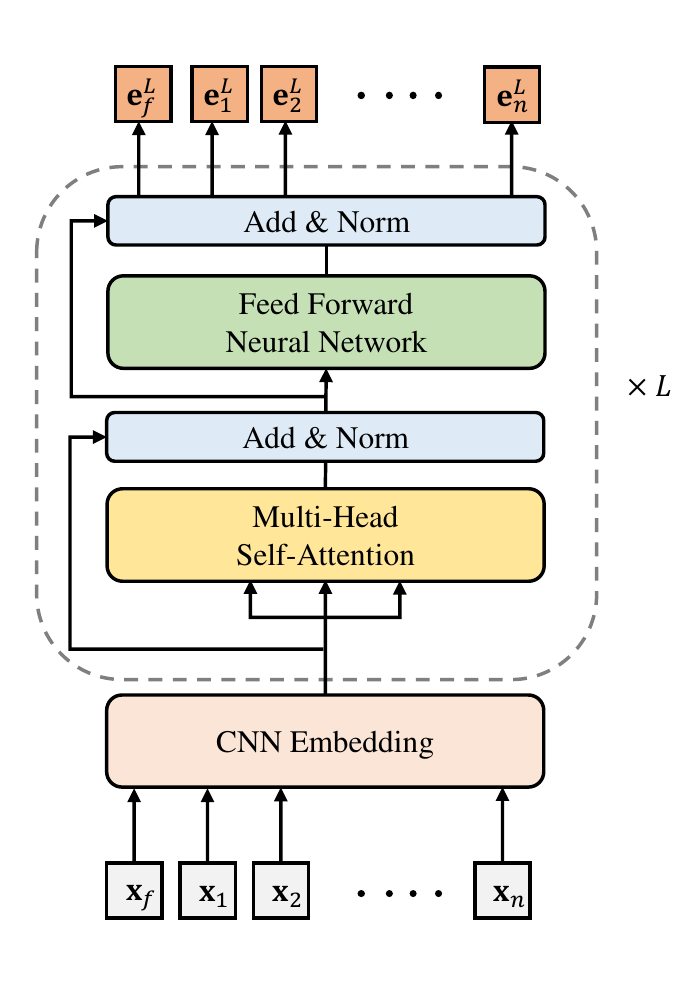}
    }
    \caption{Proposed encoder architecture with the CNN embedding layer and $L$ identical encoder layers}
    \label{fig:model_encoder}
\end{figure}

The proposed encoder is composed of a CNN embedding layer and $L$ identical encoder layers as illustrated in Figure~\ref{fig:model_encoder}. The CNN embedding layer generates embedding vectors by extracting local spatial information from the input data points, which is passed on to the subsequent encoder layers.

Each encoder layer consists of two sublayers: multi-head self-attention (MHSA) sublayer and point-wise feed forward (FF) sublayer. The MHSA sublayer  performs multi-head self-attention to capture the dependencies between each node and the point-wise FF performs non-linear activation. The residual connection \cite{he-residual} and batch normalization \cite{ioffe-bn} were incorporated between each sublayer. Similar to previous studies \cite{kool,bresson}, we use batch normalization instead of using layer normalization as it can effectively handle a large number of nodes.
%%%% Fig. 2 Encoder structure %%%%

\subsubsection{CNN embedding layer}\label{subsubsec:embedding}
We add the CNN embedding layer in the encoder for extracting spatial information from the input nodes. Let $\mathbf{X}_i^\text{$k$-NN}$ be the concatenation of feature vectors of the $i^\text{th}$ node and its $k$-nearest neighboring node feature vectors closest in distance to the $i^\text{th}$ node. 
\rev{Then, the embedding vector of the $i^\text{th}$ node in the encoder, $\x_i^\text{emb}$, is computed as the sum of node embedding of $i^\text{th}$ input nodes, $\x_i \mathbf{W}_\text{emb}$, and the convolution of $\mathbf{X}_i^\text{$k$-NN}$, which is expressed as:} %revision
%%%% Eq. 1 - CNN Embedidng %%%%
\begin{equation}
    \x^{\text{emb}}_i=\x_i\mathbf{W}_{\text{emb}} + \text{Conv}(\mathbf{X}^{\text{$k$-NN}}_i)\in\mathbb{R}^d,
\end{equation}
where $\mathbf{W}_\text{emb}\in\mathbb{R}^{2\times d}$ is a learnable parameter for node embedding.
\rev{A fixed value of kernel size $k$+1 is used to ensure that $\text{Conv}(\mathbf{X}_i^\text{$k$-NN})$ is  in the same $d$-dimensional space as node embedding, $\x_i \mathbf{W}_\text{emb}$.} %revision

\subsubsection{Encoder layer}\label{subsubsec:encoderlayer}
The encoder has $L$ identical encoder layers, and the first encoder layer takes $\x_i^\text{emb}$ from CNN embedding layer as input. Each encoder layer has two sublayers: MHSA sublayer and point-wise FF sublayer. \\

\begin{description}
\item \textbf{MHSA sublayer.} The input of the MHSA sublayer of $l^\text{th}$ encoder layer, $\E^{l-1}$, is the output of the $(l-1)^\text{th}$ encoder layer. The output of the MHSA sublayer of $l^\text{th}$ encoder layer, $\hat\E^l$, is obtained by first applying multi-head self-attention ($\MH^l$) to $\E^{l-1}$, followed by residual connection and batch normalization ($\BN^l$). The function $\MH^l$(Q, K, V) takes three inputs Q, K, V, which represent the query, key, and value vectors, respectively, to perform multi-head attention mechanism at the  $l^\text{th}$ encoder layer. The output of the MHSA sublayer, $\hat\E^l$, is formulated as follows:
%%%% Eq. 2 - Encoder MHSA %%%%
\begin{equation}
\begin{aligned}
    \hat{\E}^l = \BN^l\left(\E^{l-1} + \MH^l\left(\E^{l-1},\E^{l-1},\E^{l-1}\right)\right),
\end{aligned}
\end{equation}
where $\E^0 = \{\x_f,\x^{\text{emb}}_1,\dots,\x^{\text{emb}}_n\}\in\mathbb{R}^{(n+1)\times d}$. Here, $\E^0$, the input of the first encoder layer, is created by concatenating start token $\x_f$  with $\{\x_1^\text{emb},\dots,\x_n^\text{emb}\}$. We add $\x_f$ to create a virtual node feature vector that learns dependencies with other node features, so that the decoding can start at the best possible location \cite{bresson}. \\

\item \textbf{Point-wise FF sublayer.} The input of the point-wise FF sublayer in the $l^\text{th}$ encoder layer, which is composed of two linear projections and a ReLU activation, is $\hat\E^l$. It performs non-linear activation followed by residual connection and batch normalization and produces output $\E^l$, which is denoted as:
%%%% Eq. 3 - Encoder FF %%%%
\begin{equation}
    \E^l = \BN^l \left( \hat{\E}^l + \FF^l \left( \hat{\E}^l \right) \right),
\end{equation}
where $\FF^l$ is a FF sublayer of $l^\text{th}$ encoder layer. \\

\item \textbf{Encoder output.} Let $\e_i^L$ be the encoder output of the $i^\text{th}$ node of $L^\text{th}$ encoder layer and $f$ be the index of start token, respectively. The final encoder output of $L^\text{th}$ encoder layer, $\E^L=\{\e_f^L,\e_1^L,\dots ,\e_n^L\}$, is produced and fed into the decoder.
\end{description}
%%%%%%%% DECODER %%%%%%%%%%%
\subsection{Decoder}\label{subsec:decoder}
\begin{figure}[t!]
    \centering
    %\resizebox{\columnwidth}{!}
    {
    \includegraphics[width=9cm]{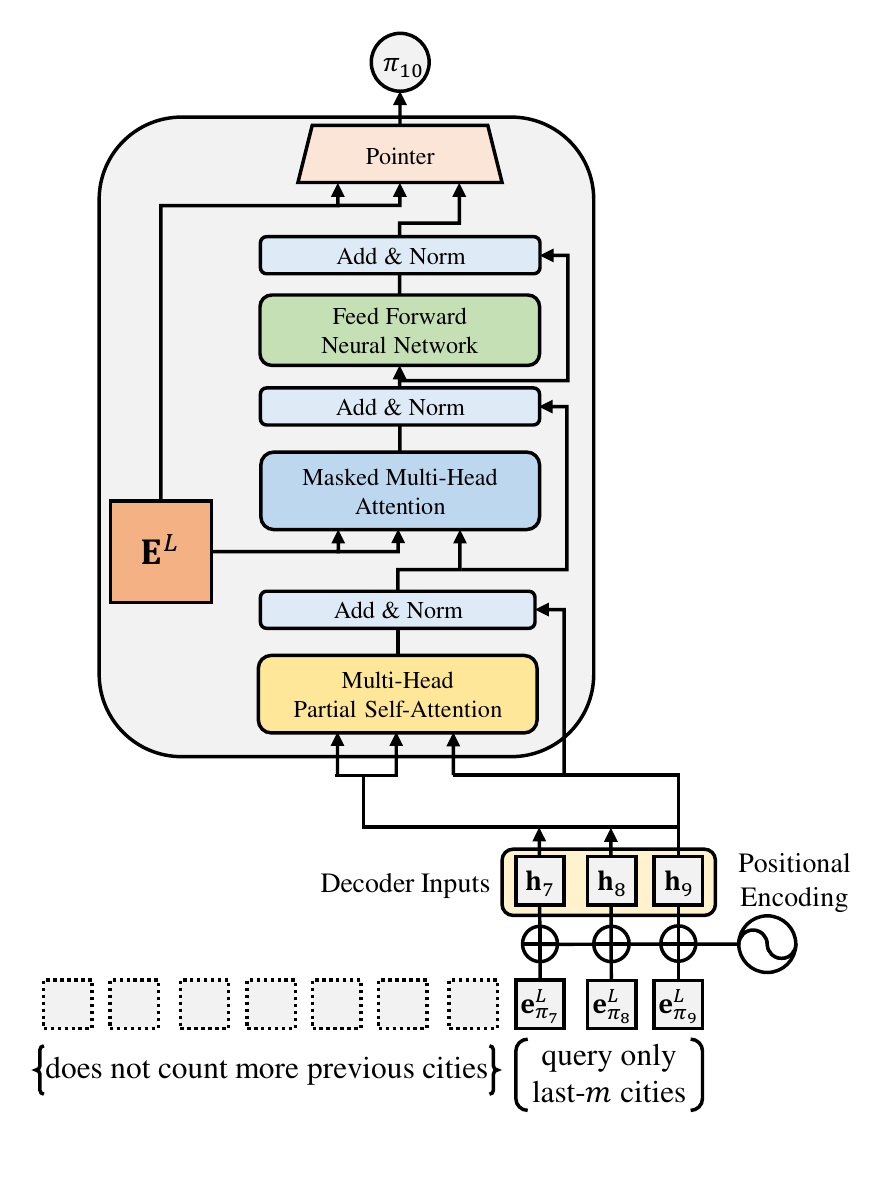}
    }
    \caption{Proposed decoder architecture. This figure shows a decoding process when the current time step $t=10$ and the number of reference vectors $m=3$}
    \label{fig:model_decoder}
\end{figure}

We perform decoding auto-regressively, one node at a time. The decoder is comprised  of four layers, each of which are followed by residual connection and layer normalization.  Figure~\ref{fig:model_decoder} illustrates an example for the decoding process when the current time step ($t$) is 10 and the number of reference vectors ($m$) is three. 

%%%% Fig. 3 Decoder structure %%%%

The first layer is multi-head partial self-attention layer (MHPSA), which extracts past information by performing attention with encoder outputs of already visited nodes in previous steps. Unlike previous works, we use fewer reference vectors for performance and computational efficiency. 
\rev{The second layer is a masked multi-head attention layer (MMHA), which performs an attention mechanism where the query is an output of the MHPSA layer. The reference vectors are the encoder outputs of unvisited nodes.}  %revision
The third layer is the point-wise FF layer, which performs linear projection and non-linear activation, similar to an point-wise FF sublayer in the encoder. The pointer layer selects the next node to visit by calculating a probability distribution over the unvisited nodes.  \\

\begin{description}
\item \textbf{MHPSA layer.} We are motivated by the fact that recently visited nodes are more relevant to the node to be selected in the current step than nodes that are visited earlier. Based on this intuitive fact,  the proposed partial self-attention performs attention only on recently visited nodes. We expect that the proposed model is able to better learn local compositionality compared with previous works based on fully-connected attention. 

MHPSA performs self-attention using the decoder input of current time step $t$, $\h_t$, as query and decoder inputs of recently visited nodes as reference vectors.
\rev{It is calculated as follows:} %revision
%%%% Eq. 4 - Decoder input %%%%
\begin{equation}
    \h_t=\mathbf{e}^L_{\pi_{t-1}} + \text{PE}_t, 
\end{equation}
where $\pi_0 = f$ and $\text{PE}_t$ denotes positional encoding at time step $t$.

The proposed partial self-attention uses decoder inputs for only the $m$ last visited nodes as reference vectors. For instance, suppose the current time step is $t$, then the decoder inputs used at time step $t-m$ to $t$ are used as reference vectors, denoted as $\mathbf{H}_t=\{\h_{t-m},\dots,\h_t\}\in\mathbb{R}^{m\times d}$ . Consequently, memory usage and computation time are significantly reduced. The output of MHPSA layer, $\hat\h_t$, is calculated as follows:
%%%% Eq. 5 - Decoder MHPSA %%%%
\begin{equation}
    \hat\h_t = \LN \left( \h_t + \MH^{L+1} \left( \h_t,\mathbf{H}_t,\mathbf{H}_t \right) \right), 
\end{equation}
where  $\LN$ refers to layer normalization. 
\\
\item \textbf{MMHA layer.} Masked multi-head attention layer performs attention mechanism using $\hat\h_t$ as query and $\E^L$ as reference vectors. We use the same masked multi-head attention layer as \cite{bresson}. 
Let $\zeta_t\in\mathbb{R}^{n+1}$ be the mask where the value is one for unvisited and zero for visited nodes to the attention weight, respectively. Then, the output from the MMHA layer is calculated as follows:
%%%% Eq. 6 - Decoder MMHSA %%%%
\begin{equation}
    \tilde\h_t = \LN \left( \hat\h_t + \text{MMH}^{L+1} \left(\hat\h_t,\E^L,\E^L,\zeta_ t\right) \right),
\end{equation}
where $\text{MMH}(Q, K, V, \zeta)$ is a modified function of $\MH(Q, K, V)$, which takes an additional input mask $\zeta$ and replaces Attention function \cite{vaswani} with $\text{MaskedAttention}$ function formulated as follows:

\begin{equation*}
%\resizebox{\columnwidth}{!}{
\text{MaskedAttention}(Q, K, V,\zeta) = \text{softmax}\left( \frac{QK^\intercal}{\sqrt{d_k}}\odot\zeta  \right) V,
%}
\end{equation*}
where $\odot$ denotes element-wise product operation. 
\\
\item \textbf{Point-wise FF layer.} The input of the point-wise FF layer is $\tilde\h_t$ and the output is $\bar\h_t$, which is denoted as:
%%%% Eq. 7 - Decoder FF %%%%
\begin{equation}
    \bar\h_t = \LN \left( \tilde\h_t + \FF^{L+1} \left( \tilde\h_t \right) \right).
\end{equation} 
\\
\item \textbf{Pointer layer.} The goal of the pointer layer is to compute a distribution over unvisited nodes. We perform single-head attention by using $\bar\h_t$ as query and $\E^L$  as reference vectors. We use attention weights as probability distributions, $p_t$, that determine the  next node to visit.
\rev{Masking is used to avoid already visited nodes.} %revision
Then, $p_t$ can be computed as: 
%%%% Eq. 8 - Pointer layer %%%%
\begin{equation}
    p_t = \text{softmax} \left( c\cdot \tanh \left( \frac{qK^\intercal}{\sqrt{d}} \right) \odot\zeta_t \right),
\end{equation}
where $q$ and $K$ are query and reference vectors of single head attention and $c$ is a
hyperparameter that controls the range of the logits~\cite{bello}.

During the training phase, the decoder considers $p_t$ as a categorical distribution for sampling node indices,  and the node index with the highest probability is selected during the inference phase. This process is repeated $n$ times, resulting in a  node indices sequence $\tour=\{\pi_1,\dots,\pi_n\}$, which is the final output of the model.
\end{description}

%%%%%%%% REINFORCEMENT LEARNING %%%%%%%
\subsection{Model training based on reinforcement learning}\label{subsec:rl}
\rev{We trained our model via reinforcement learning, utilizing tour length as a negative reward.}  %revision
The loss function is the average tour length. Let $\theta$ be training model parameters. Then, $p(\tour;\theta)$ is the probability that the model generates $\tour$, which can be defined as:
%%%% Eq. 9 - Chain rule %%%%
\begin{equation}
    p(\tour;\theta) = \prod^n_{t=1} p(\pi_t|\pi_1,\dots,\pi_{t-1};\theta),
\end{equation}
where $p(\pi_t |\pi_1,\dots ,\pi_{t-1};\theta)$ is the probability that $\pi_t$ is chosen from $p_t$ at time step $t$. We use REINFORCE algorithm to update $\theta$. A duplicate version of $\theta$, $\theta_b$, is used as a baseline. We denote $\tour'$ an index sequence that the model parameterized by $\theta_b$ generates in a greedy way. Let $\ell(\tour)$ and $\ell(\tour')$ be the total tour length of node sequences of training and baseline models, respectively. Then, a REINFORCE loss $L(\theta)$ is optimized by gradient descent methods using the REINFORCE algorithm:
%%%% Eq. 10 - Loss %%%%
\begin{equation}
    L(\theta) = \mathbb{E}_{\tour\sim p(\tour;\theta)} \left[ \ell \left( \tour \right) -\ell \left( \tour' \right) \right].
\end{equation}
The gradient of $L(\theta)$ is computed as:
\begin{equation}
    \nabla_\theta L(\theta) \approx\sum_{\tour} \left( \ell \left( \tour \right) - \ell \left( \tour' \right) \right) \nabla_\theta \log p(\tour;\theta).
\end{equation}
\rev{We optimize the training model using $\nabla_\theta L(\theta)$ during one epoch.} %revision
 When one epoch ends, we compare the average tour length of training and baseline models. We copy $\theta$ to $\theta_b$ if the average tour length of the training model is shorter than that of the baseline  model.

%%%%%%%%%%%%%%%%%%%%%%%%%
%%%%% Experiment %%%%%%%%
%%%%%%%%%%%%%%%%%%%%%%%%%
\section{Experiment}\label{sec:experiment}
\subsection{Datasets}\label{subsec:datasets}
\begin{description}
\item \textbf{Random dataset.} \rev{We assume a 2D planar symmetric TSP, where the distance between two cities in opposite directions are equal.} %revision
For model training and validation, we use a randomly generated data from a uniform distribution on the fly in $\left [ 0, 1 \right ] \times \left [ 0, 1 \right ]$. We generated 10,000 test instances. 
\rev{We trained and tested on TSP problems from $n = 50$ (TSP50) to $200$ nodes (TSP200).} %revision
 The output tour using Concorde \cite{concorde} was considered to have exact solutions and labels of test instances. \\

\item \textbf{TSPLIB dataset.} TSPLIB \cite{tsplib} is a widely-used benchmark dataset for evaluating the performance of TSP solvers on a variety of real-world data with varying distributions. We choose ten 2D-Euclidean problem instances from TSPLIB, which are considered relatively difficult. Let $N$ and $A$ be the number of nodes and the square area covered by the nodes, respectively \cite{sultana}. Then, we use a critical parameter value, $\frac{l}{\sqrt{N\cdot A}}$, to evaluate the difficulty level of TSPLIB where $l$ is the optimal tour length \cite{hardcase1,hardcase2}. A critical parameter value close to $0.75$ indicates a higher difficulty level. We normalize each TSPLIB instance such that all TSPLIB instances are in $\left [ 0, 1 \right ] \times \left [ 0, 1 \right ]$.
\end{description}
\subsection{Hyperparameters and decoding strategy}\label{subsec:hparams}
\begin{description}
\item \textbf{Hyperparameters.} We did not tune the hyperparameters of the proposed model. We use the same hyperparameters for all TSP problem sizes. The proposed model has six encoder layers ($L=6$) with $k=10$. The kernel size of CNN embedding layers is set to 11. The model has 128 hidden dimensions ($d=128$) and the hidden dimension of each point-wise FF layer (also sublayer) and is set to 512. The model has eight attention heads. The logit range clipping value $c$ in the decoder is set to 10.

For model training, we use an Adam optimizer with a fixed learning rate of 0.0001. A batch size of 512 is used. We train for 100 epochs using training data but further increasing the number of epochs could improve the model's performance. Our experiments were conducted on an AMD EPYC 7513 32-Core Processor and a single Nvidia A6000 GPU. \\

\item \textbf{Decoding strategy.} At test time, we employ both greedy and beam search for decoding. Beam search \cite{beamsearch} is a breadth-first search strategy that considers top-B cases in every decoding time step and chooses the best solution at the end of the decoding. We set the beam width (B) as 2,500 in order to compare our results with other SOTA models \cite{bresson} that use beam width of 2,500.
\end{description}
%%%%%%%%%%%%%%%%%%%%%%%%%%%%%%%%%%%%%%%%%%%%%
%%%%%%%%%  Table 1 - Model Architecture Comparison  % revision02-26
%%%%%%%%%%%%%%%%%%%%%%%%%%%%%%%%%%%%%%%%%%%%%
\begin{table*}[h!]
\centering
\caption{\rev{Comparison of Transformer-based models for TSP: Kool et al. \cite{kool}, TSP Transformer \cite{bresson}, Tspformer \cite{memory-eff}, and our model. Here, MHSA and MMHA denote  Multi-Head Self-Attention and Masked Multi-Head Attention, respectively. MHSA in the decoder was not used in Kool et al. \cite{kool}}}
\label{tab:model-comp}
\resizebox{\textwidth}{!}{%
\begin{tabular}{lccccccc}
\hline
\multirow{3}{*}{Model} & \multicolumn{3}{c}{Encoder}                             & \multicolumn{4}{c}{Decoder}                                                                                                                                                                                                                     \\ \cline{2-8} 
                       & \multirow{2}{*}{Embedding} & \multicolumn{2}{c}{MHSA}   & \multicolumn{2}{c}{MHSA}                                                     & \multicolumn{2}{c}{MMHA}                                                                                                                                         \\ \cline{3-8} 
                       &                            & Query         & Key        & Query     & Key                                                              & Query                                                                                          & Key                                                             \\ \hline
Kool et al. \cite{kool}                   & Linear                     & Every node    & Every node & --        & --                                                               & \multicolumn{1}{l}{\begin{tabular}[c]{@{}l@{}}First node\\ + last node\\ + graph\end{tabular}} & \begin{tabular}[c]{@{}c@{}}Every \\ unvisited node\end{tabular} \\
TSP Transformer \cite{bresson}        & Linear                     & Every node    & Every node & Last node & \begin{tabular}[c]{@{}c@{}}Every\\ visited node\end{tabular}     & Last node                                                                                      & \begin{tabular}[c]{@{}c@{}}Every \\ unvisited node\end{tabular} \\
Tspformer \cite{memory-eff}              & Linear                     & Sampled nodes & Every node & --        & --                                                               & Last node                                                                                      & \begin{tabular}[c]{@{}c@{}}Every \\ unvisited node\end{tabular} \\
Ours                   & CNN                        & Every node    & Every node & Last node & \begin{tabular}[c]{@{}c@{}}Last-$m$\\ visited nodes\end{tabular} & Last node                                                                                      & \begin{tabular}[c]{@{}c@{}}Every\\ unvisited node\end{tabular}  \\ \hline
\end{tabular}%
}
\end{table*}
%%%%%%%%%%%%%%%%%%%%%%%%%%%%%%%%%%%%%%%%%%%%%
%%%%%%%%%%%%%%%%%%%%%%%%%%%%%%%%%%%%%%%%%%%%%

\subsection{List of Experiments}\label{subsec:list-experiment}
In our study, we conducted four experiments to evaluate the proposed model's performance. The third experiment used both the random dataset and the TSPLIB dataset, while the other experiments only used the random dataset.
\rev{For Experiments 1 and 2, we employed greedy decoders to find optimal tours.}  %revision
 \\
\begin{description}
\item \textbf{Experiment 1.} We test whether the proposed partial self-attention in the decoder is more effective in improving the performance of the model compared with the existing fully-connected self-attention in the standard Transformer decoder. 
\rev{We test by decreasing the number of reference vectors ($m$) from 200 to 5.} %revision
\\

\item \textbf{Experiment 2.} We remove the CNN embedding layer in our CNN-Transformer model and test whether it is effective in extracting local spatial information and produces better output performance by performing an ablation study. \\

\item \textbf{Experiment 3.} \rev{We compare the proposed model with other solvers, including the optimal solver (Concorde \cite{concorde}), heuristic solvers such as 2-opt search \cite{two-opt}, Monte Carlo Tree Search (MCTS) \cite{kocsis-mcts}, Google OR-Tools \cite{ortools}, and other SOTA Transformer-based models for TSP \cite{kool,bresson,memory-eff,h-tsp}.} %revision
 Table \ref{tab:model-comp} summarizes the main difference between the proposed model with other Transformer-based models. Here, `node' means the hidden feature vector for a particular node generated by the encoder or MHSA layer of each model and `graph' is the average value of the hidden feature vectors of all nodes. \\

\item \textbf{Experiment 4.} We compare the computational complexity between the proposed CNN-Transformer model with other Transformer-based models.
\end{description}

\subsection{Metrics}\label{subsec:metrics}
The performance of the model was evaluated using the following metrics. \\

\begin{description}
\item \textbf{Average predicted tour length.} Let $\hat l_i^\text{TSP}$ be the predicted tour length of the $i^\text{th}$ instance. Then, the average tour length is computed as $\frac{1}{n}\sum_{i=1}^n \hat l_i^\text{TSP}$ , where $n$ is the number of total test instances. Here, we set $n$ as 10,000. \\

\item \textbf{Optimality gap.} The optimality gap is computed as the average percentage of the predicted tour length to the optimal solution, which is computed as $\frac{1}{n}\sum_{i=1}^n \left( \frac{\hat l_i^\text{TSP}}{l_i^\text{TSP}}-1 \right)$, where $n$ is  the number of total test instances. Here, $l_i^\text{TSP}$ is the optimal solution of the $i^\text{th}$ instance produced by Concorde \cite{concorde}. We set $n$ as 10,000. \\

\item \textbf{Training time.} \rev{The training time is measured as the time taken to train 10,000 instances with batch size of 512.}  \\ %revision

\item \textbf{Inference time.} \rev{We report the inference time taken to solve the test set of 100 instances of TSP100 and TSP200.}  %revision
Beam width (B) was set to 2,500 and batch size was set to one due to the limitation of memory capacity. \\

\item \textbf{GPU Memory Usage.} Maximal memory usage capacitated by the training process is measured. \\
\end{description}

%%%%%%%%%%%%%%%%%%%%%%%%%
%%%%% RESULTS %%%%%%%%%%%
%%%%%%%%%%%%%%%%%%%%%%%%%
\section{Results}\label{sec:results}

\subsection{Experiment 1}\label{subsec:expr1}

Table~\ref{tab:m-comp} presents  the average tour length and the optimality gap of our model with various $m$ values. 
\rev{The proposed model showed the best performance when $m$ was $100$ for TSP150 and $20$ for TSP200. Our experimental results show that the optimal $m$ value changes with respect to problem size.} %revision
 For TSP100, the proposed model using partial self-attention with $m=5$ achieved an optimality gap of 2.83\%, but the proposed model using the fully-connected attention, i.e., $m=100$, produced an optimality gap of 3.10\%.

%%%%%%%%%%%%%%%%%%%%%%%%%%%%%%%%%%%%%%%%%%%%%
%%%%%%%%%  Table 2 - varied m
%%%%%%%%%%%%%%%%%%%%%%%%%%%%%%%%%%%%%%%%%%%%%
\begin{table*}[ht]
\centering\caption{\rev{Average tour length (Obj.) and the optimality gap (Gap) of our model with various $m$ (number of reference vectors) values for various TSP instance sizes using the random datasets. The greedy decoder is used for this experiment}}
\label{tab:m-comp}
\begin{tabular}{lrrrrrrrr}
\hline
    & \multicolumn{2}{c}{TSP50}      & \multicolumn{2}{c}{TSP100}     & \multicolumn{2}{c}{TSP150}     & \multicolumn{2}{c}{TSP200}       \\
$m$   & Obj.       & Gap           & Obj.       & Gap           & Obj.       & Gap           & Obj.        & Gap            \\ \hline
200 & --     & --   & --   & --   & --     & --   & 12.417   & 15.97\%          \\
150 & --     & --    & --    & --    & 9.793     & 4.60\%    & 12.634    & 18.00\%             \\
100 & --     & --       & 8.005          & 3.10\%           & \textbf{9.764} & \textbf{4.29}\% & 12.344          & 15.28\%          \\
50  & 5.750           & 1.06\%          & 7.993          & 2.94\%          & 9.778          & 4.44\%          & 12.323          & 15.09\%          \\
20  & 5.748          & 1.01\%          & 7.988          & 2.87\%          & 9.779          & 4.45\%          & \textbf{12.004} & \textbf{12.11}\% \\
5   & \textbf{5.745} & \textbf{0.97}\% & \textbf{7.985} & \textbf{2.83}\% & 9.773          & 4.38\%          & 12.353          & 15.37\%          \\ \hline
\end{tabular}
\end{table*}
%%%%%%%%%%%%%%%%%%%%%%%%%%%%%%%%%%%%%%%%%%%%%
%%%%%%%%%%%%%%%%%%%%%%%%%%%%%%%%%%%%%%%%%%%%%

%%%%%%%%%%%%%%%%%%%%%%%%%%%%%%%%%%%%%%%%%%%%%
%%%%%%%%% Table 3 - Ablation study  % revision02-26
%%%%%%%%%%%%%%%%%%%%%%%%%%%%%%%%%%%%%%%%%%%%%
\begin{table*}[ht]
\centering\caption{\rev{Ablation study. Average tour length (Obj.) and the optimality gap (Gap) of our model without CNN and with CNN for various TSP instances using the random dataset. The greedy decoder is used for this experiment}}
\label{tab:ablation}
\resizebox{\textwidth}{!}
{
\begin{tabular}{lrrrrrrrr}
\hline
                   & \multicolumn{2}{c}{TSP50}                 & \multicolumn{2}{c}{TSP100}     & \multicolumn{2}{c}{TSP150}     & \multicolumn{2}{c}{TSP200}       \\
                   & Obj.                  & Gap           & Obj.       & Gap           & Obj.       & Gap           & Obj.        & Gap            \\ \hline
\begin{tabular}[c]{@{}l@{}}Ours\\ (without CNN)\end{tabular} & \multicolumn{1}{c}{5.754} & 1.12\%          & 8.004          & 3.08\%          & 9.811          & 4.79\%          & 12.306          & 14.93\%          \\
Ours               & \textbf{5.745}            & \textbf{0.97}\% & \textbf{7.985} & \textbf{2.83}\% & \textbf{9.764} & \textbf{4.29}\% & \textbf{12.004} & \textbf{12.11}\% \\ \hline
\end{tabular}
}
\end{table*}
%%%%%%%%%%%%%%%%%%%%%%%%%%%%%%%%%%%%%%%%%%%%%
%%%%%%%%%%%%%%%%%%%%%%%%%%%%%%%%%%%%%%%%%%%%%

\subsection{Experiment 2}\label{subsec:expr2}

Table \ref{tab:ablation} presents the results of the ablation study. Both models use greedy decoders. 
\rev{Our results show that for all TSP problem sizes, the proposed model outperforms the proposed model without the CNN embedding layer.} %revision
We observed that removing the CNN embedding layer degrades the overall performance. Therefore, we can conclude that the CNN embedding layer in our model is effective in extracting spatial features and improving the output performance.

\subsection{Experiment 3}\label{subsec:expr3}

\begin{figure*}[t!]
    \centering
    \includegraphics[width=\textwidth]{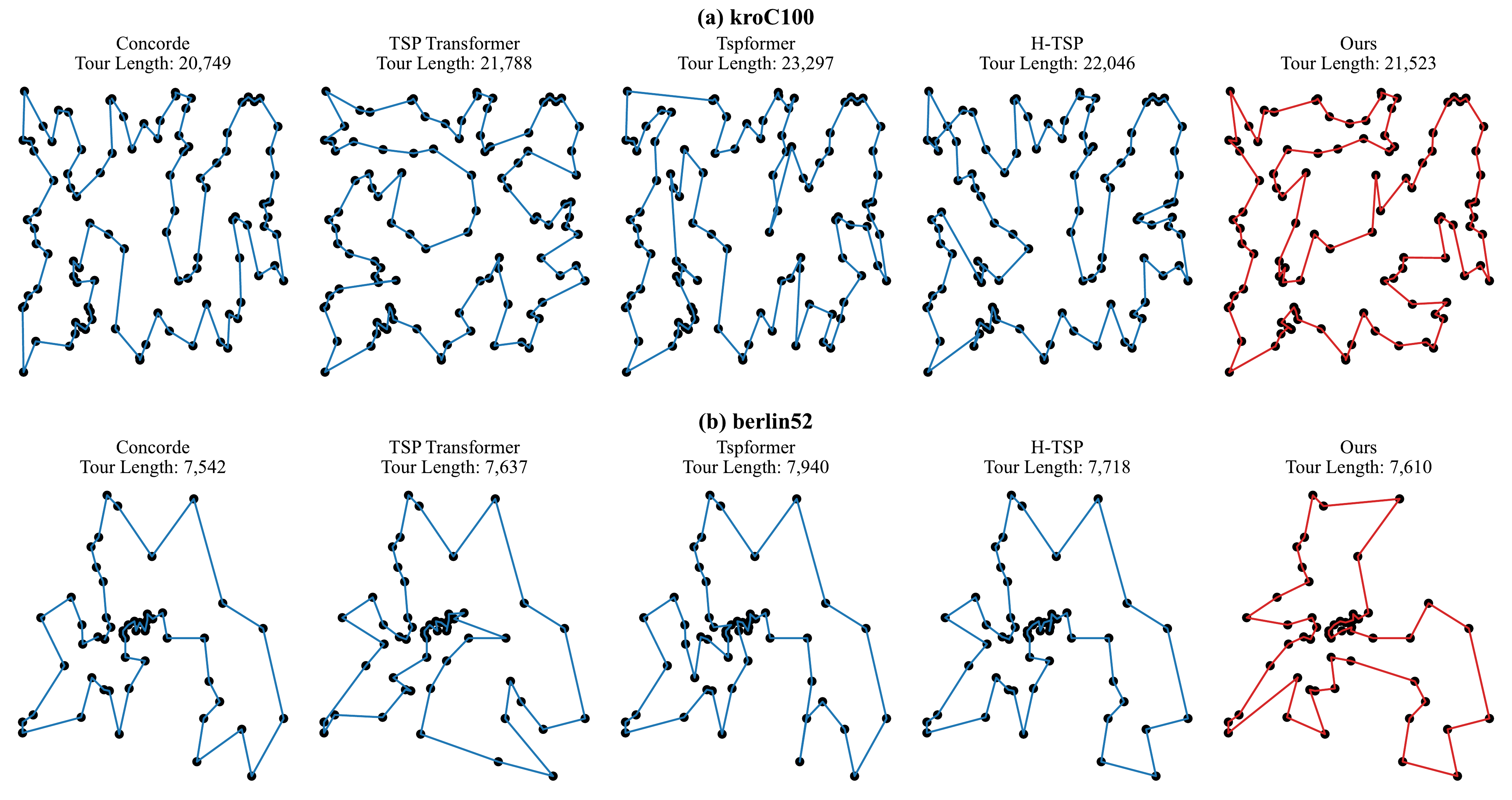}
    \caption{\rev{Output tours of Concorde \cite{concorde}, TSP Transformer \cite{bresson}, Tspformer \cite{memory-eff}, H-TSP \cite{h-tsp}, and our model for (a) kroC100 and (b) berlin52 using the TSPLIB dataset}}
    \label{fig:tsplib_output}
\end{figure*}

\begin{description}
\item \textbf{Random dataset.} Table \ref{tab:random-tsp} compares the performance of the proposed model with other SOTA Transformer-based models. We present the results by dividing the table into three sections: exact solver, heuristics, and deep learning models. Concorde \cite{concorde}, which is known to produce optimal results, shows the best performance. Among deep learning models, the proposed model has the best performance both for greedy and beam search decoding. 
\rev{Our experimental results showed that our model significantly outperformed MCTS \cite{kocsis-mcts} and the 2-opt search method \cite{two-opt} in performance and generalization.} %revision
 For TSP50, the proposed model is only behind by a 0.1\% optimality gap  compared with the exact solver, Concorde. 
\rev{Our model using beam search reduces the optimality gap from 0.11\% to 0.10\% for TSP50 and 1.26\% to 1.11\% for TSP100 compared with the TSP Transformer.} %revision
 
\rev{The proposed model shows competitive performance compared with H-TSP for random datasets. The proposed model showed better performance for TSP50 and TSP100 compared with the H-TSP, whereas the H-TSP showed better performance for TSP 200. The reason is that the H-TSP is known as a specialized model for large-scale TSP, showing outstanding performance in such instances.} %revision
  \\

\item \textbf{TSPLIB dataset.} \rev{Table \ref{tab:tsplib} presents the output performances for our model and other Transformer-based models on various TSPLIB instances.} %revision
\rev{All models were trained on TSP50.} %revision
 We observe that the proposed model gives  consistent results for various TSPLIB instances and outperforms TSP Transformer in most TSPLIB instances. Therefore, the proposed model using partial self-attention is also effective in real-world datasets.
\rev{The proposed model outperforms H-TSP for all TSPLIB instances and outperforms TSP Transformer for all TSPLIB instances except for one case, i.e., rd100.} %revision

Two most difficult TSPLIB instances are the kroC100 and the berlin52 instances. The berlin52 is known to be a hard TSP instance because many nodes are highly constrained in very small regions. 
\rev{For very hard real-world instances, the proposed model showed the best performance among compared models.} %revision
\rev{Figure \ref{fig:tsplib_output} displays the output of Concorde and various models on kroC100 and berlin52.} %revision 
For the kroC100, the optimal tour length of Concorde is 20,749.
\rev{The tour length of our model is 21,523, while that of the TSP Transformer model is 21,788. For the berlin52, the tour length of our model is 7,610, but the tour length of the TSP Transformer model is 7,637.} %revision 
 \\
\end{description}

\subsection{Experiment 4}\label{subsec:expr4}

%%%%%%%%%%%%%%%%%%%%%%%%%%%%%%%%%%%%%%%%%%%%%
%%%% Table 4 - Random TSP
%%%%%%%%%%%%%%%%%%%%%%%%%%%%%%%%%%%%%%%%%%%%%
\begin{table*}[t]
\centering
\caption{\rev{Average tour length (Obj.) and optimality gap (Gap) of various models and solvers for various TSP instance sizes using the random datasets. Results with * are reported from other papers. In the $type$ column, H: Heuristic, RL: Reinforcement learning, G: Greedy, S: Sampling and BS: Beam search}}
\label{tab:random-tsp}
\resizebox{\textwidth}{!}
{
\begin{tabular}{lcrrrrrrrr}
\hline
                                 &                            & \multicolumn{2}{c}{TSP50}       & \multicolumn{2}{c}{TSP100}      & \multicolumn{2}{c}{TSP150}      & \multicolumn{2}{c}{TSP200}       \\
Method                           & Type                       & Obj.       & Gap            & Obj.       & Gap            & Obj.       & Gap            & Obj.        & Gap            \\ \hline
Concorde \cite{concorde}                         & \begin{tabular}[c]{@{}l@{}}Exact\\ solver\end{tabular}               & 5.689          & 0.00\%          & 7.764          & 0.00\%          & 9.363          & 0.00\%          & 10.708          & 0.00\%          \\ \hline
MCTS \cite{kocsis-mcts}          &  \multirow{3}{*}{H}                          & 7.323           & 28.72\%         & 12.092          & 55.74\%         & 17.230          & 84.02\%         & 22.475           & 109.89\%        \\
2-opt search \cite{two-opt}                     &  & 6.451           & 13.39\%         & 9.180           & 18.24\%         & 11.277          & 20.44\%         & 13.084           & 22.19\%         \\
OR Tools \cite{ortools}                         &                            & 5.863           & 3.06\%          & 8.094           & 4.25\%          & 9.826           & 4.94\%          & 11.280           & 5.34\%          \\ \hline
\multirow{2}{*}{Kool et al. \cite{kool}}     & RL, G                      & 5.795           & 1.86\%          & 8.107           & 4.42\%          & 9.938           & 6.14\%          & 13.246           & 23.70\%         \\
                                 & RL, S                      & 5.727           & 0.67\%          & 7.941           & 2.28\%          & 9.709           & 3.70\%          & 12.227           & 14.19\%         \\
\multirow{2}{*}{TSP Transformer \cite{bresson}} & RL, G                      & 5.750          & 1.05\%          & 8.015          & 3.22\%          & 9.814          & 4.83\%          & 12.571          & 17.40\%          \\
                                 & RL, BS                     & 5.695          & 0.11\%          & 7.863          & 1.26\%          & 9.616          & 2.71\%          & 12.135          & 13.33\%         \\
\multirow{2}{*}{Tspformer \cite{memory-eff}}       & RL, G                      & 5.874          & 3.25\%          & 8.175          & 5.29\%          & 9.958          & 6.35\%          & 13.371          & 24.87\%         \\
                                 & RL, BS                     & 5.726          & 0.65\%          & 7.953          & 2.43\%          & 9.706          & 3.66\%          & 12.865          & 20.14\%         \\
H-TSP \cite{h-tsp}                            & RL, G                       & 5.748          & 1.04\%          & 7.889          & 1.61\%          & \textbf{9.574} & \textbf{2.25}\% & \textbf{10.990} & \textbf{2.63}\% \\
\multirow{2}{*}{Ours}            & RL, G                      & 5.745          & 0.97\%          & 7.985          & 2.83\%          & 9.764          & 4.29\%          & 12.004          & 12.11\%         \\
                                 & RL, BS                     & \textbf{5.695} & \textbf{0.10}\% & \textbf{7.851} & \textbf{1.11}\% & 9.589          & 2.42\%          & 11.637          & 8.68\%          \\ \hline
\end{tabular}
}
\end{table*}
%%%%%%%%%%%%%%%%%%%%%%%%%%%%%%%%%%%%%%%%%%%%%
%%%%%%%%%%%%%%%%%%%%%%%%%%%%%%%%%%%%%%%%%%%%%

%%%%%%%%%%%%%%%%%%%%%%%%%%%%%%%%%%%%%%%%%%%%%
%%%% Table 5 - TSPLIB
%%%%%%%%%%%%%%%%%%%%%%%%%%%%%%%%%%%%%%%%%%%%%
\begin{table*}[t]
\centering
\caption{\rev{Tour length (Obj.) and optimality gap (Gap) for the TSPLIB instances. The critical parameters $\frac{l}{\sqrt{N\cdot A}}$ close to 0.75 indicate harder TSP instances. The greedy decoder is used for this experiment}}
\label{tab:tsplib}
\resizebox{\textwidth}{!}{%
\begin{tabular}{lrrrrrrrrrrrr}
\hline
\multirow{2}{*}{Problem} & \multirow{2}{*}{$\frac{l}{\sqrt{N\cdot A}}$} & \multicolumn{1}{c}{\multirow{2}{*}{Concorde \cite{concorde}}} & \multicolumn{2}{c}{Kool et al. \cite{kool}}  & \multicolumn{2}{c}{TSP Transformer \cite{bresson}} & \multicolumn{2}{c}{Tspformer \cite{memory-eff}} & \multicolumn{2}{c}{H-TSP \cite{h-tsp}} & \multicolumn{2}{c}{Ours}          \\
                         &                                             & \multicolumn{1}{c}{}                          & Obj.       & Gap             & Obj.          & Gap             & Obj.       & Gap          & Obj.     & Gap        & Obj.        & Gap             \\ \hline
kroC100                  & 0.75                                        & 20,749                                        & 21,565         & 3.93\%          & 21,788            & 5.01\%          & 23,297         & 12.28\%      & 22,046       & 6.25\%     & \textbf{21,523} & \textbf{3.73\%} \\
berlin52                 & 0.74                                        & 7,542                                         & 8,017          & 6.30\%           & 7,637             & 1.26\%          & 7,940          & 5.28\%       & 7,718        & 2.33\%     & \textbf{7,610}  & \textbf{0.90\%} \\
kroA100                  & 0.77                                        & 21,282                                        & 23,078         & 8.44\%          & 21,747            & 2.18\%          & 24,008         & 12.81\%      & 22,758       & 6.94\%     & \textbf{21,620} & \textbf{1.59\%} \\
ch150                    & 0.78                                        & 6,528                                         & 7,242          & 10.94\%         & 7,390             & 13.20\%          & 7,274          & 11.43\%      & 7,268        & 11.34\%    & \textbf{7,050}  & \textbf{8.00\%} \\
ch130                    & 0.78                                        & 6,110                                         & \textbf{6,549} & \textbf{7.18\%} & 6,569             & 7.51\%          & 6,808          & 11.42\%      & 6,601        & 8.04\%     & 6,552           & 7.23\%          \\
rd100                    & 0.81                                        & 7,910                                         & 8,441          & 6.71\%          & 8,078             & 2.12\%          & 8,549          & 8.08\%       & 8,481        & 7.22\%     & \textbf{8,044}  & \textbf{1.69\%} \\
st70                     & 0.86                                        & 675                                           & 698            & 3.41\%          & 710               & 5.19\%          & 701            & 3.85\%       & 698          & 3.41\%     & \textbf{676}    & \textbf{0.15\%} \\
eil101                   & 0.98                                        & 629                                           & \textbf{665}   & \textbf{5.72\%} & 681               & 8.27\%          & 690            & 9.70\%        & 677          & 7.63\%     & 668             & 6.20\%          \\
eil76                    & 1.03                                        & 538                                           & \textbf{560}   & \textbf{4.09\%} & 565               & 5.02\%          & 592            & 10.04\%      & 569          & 5.76\%     & 564             & 4.83\%          \\
eil51                    & 1.05                                        & 426                                           & 439            & 3.05\%          & 438               & 2.82\%          & 435            & 2.11\%       & 438          & 2.82\%     & \textbf{429}    & \textbf{0.70\%} \\ \hline
\end{tabular}%
}
\end{table*}
%%%%%%%%%%%%%%%%%%%%%%%%%%%%%%%%%%%%%%%%%%%%%
%%%%%%%%%%%%%%%%%%%%%%%%%%%%%%%%%%%%%%%%%%%%%

%%%%% Fig. 4 TSPLIB output %%%%%

\rev{Table \ref{tab:comp-comp} summarizes the overall model complexity and runtimes of various Transformer-based models for TSP100 and TSP200. The proposed model shows competitive GPU memory usage, training time, and inference time compared to other models. Our results show that the computational complexity of the H-TSP is highest among the compared models. It requires an additional warm-up stage for the lower-level model by pre-training and undergoes a rather complex training process. Therefore, it has a drawback of demanding substantial computing resources and significant training time.} 

\rev{The H-TSP has 3.7 times more model parameters and the training time was approximately 68 times longer than our model for TSP200. The proposed model has more model parameters than the TSP Transformer, as we added the CNN embedding layer in our model. However, our model consumes 17.8\% and 19.8\% less GPU memory for TSP100 and TSP200, respectively. Compared to the Tspformer model proposed by Yang et al. \cite{memory-eff}, our model exhibits slightly higher computational complexity.} %revision (with above paragraph)

% The main strength of our model lies in its reduced GPU memory usage due to partial self-attention in the decoder. We remove redundant attention connections by applying partial self-attention in the decoder, which reduces inference time. %revision (with above paragraph) %

%%%%%%%%%%%%%%%%%%%%%%%%%%%%%%%%%%%%%%%%%%%%%
%%%%% Table 6 - Computational complexity
%%%%%%%%%%%%%%%%%%%%%%%%%%%%%%%%%%%%%%%%%%%%%
\begin{table*}[t!]
\centering
\caption{\rev{Comparison of model parameters, GPU memory usage, and training (T) time for 10,000 TSP instances and inference (I) time for 100 TSP instances of TSP100 and TSP200. G: Greedy and BS: Beam search}}
\label{tab:comp-comp}
\resizebox{\textwidth}{!}{
\begin{tabular}{llrrrrrr}
\hline
\begin{tabular}[c]{@{}l@{}}Problem\\size\end{tabular}            & Model           & \#Params & \begin{tabular}[r]{@{}r@{}}GPU Memory\\ Usage (GB)\end{tabular} & \begin{tabular}[r]{@{}r@{}}Pre-training\\ time\end{tabular} & T time   & I time (G) & I time (BS) \\ \hline
\multirow{4}{*}{TSP100} & TSP Transformer \cite{bresson} & 1.41M   & 16.59                                                           & --                                                          & 18.27s   & 10.05s & 1m15s \\
                        & Tspformer \cite{memory-eff}       & 1.08M   & 12.46                                                           & --                                                          & 15.66s   & 8.55s & 1m04s  \\
                        & H-TSP \cite{h-tsp}          & 5.3M    & 20.12                                                           & 1m12s                                                       & 21m37s   & 8.99s & --  \\
                        & Ours            & 1.43M   & 13.63                                                           & --                                                          & 18.08s   & 11.47s & 51.70s \\ \hline
\multirow{4}{*}{TSP200} & TSP Transformer \cite{bresson} & 1.41M   & 29.28                                                           & --                                                          & 1m9s     & 19.67s & 4m50s \\
                        & Tspformer \cite{memory-eff}       & 1.08M   & 20.85                                                           & --                                                          & 57.64s   & 16.75s & 4m04s \\
                        & H-TSP \cite{h-tsp}           & 5.3M    & 22.04                                                           & 5m40s                                                       & 76m43s & 17.95s & -- \\
                        & Ours            & 1.43M   & 23.48                                                           & --                                                          & 1m8s     & 23.09s & 3m22s \\ \hline
\end{tabular}
}
\end{table*}
%%%%%%%%%%%%%%%%%%%%%%%%%%%%%%%%%%%%%%%%%%%%%
%%%%%%%%%%%%%%%%%%%%%%%%%%%%%%%%%%%%%%%%%%%%%

\section{Discussion}\label{sec:discussion}
Recent studies tried several heuristic search algorithms such as Monte Carlo Tree Search \cite{xing-gnn-mcts,xing-dnn-mcts} or 2-opt search \cite{deudon,wu} to further enhance the quality of TSP solutions. The proposed model uses beam search decoding techniques, but other heuristic search algorithms for TSP can be combined to further enhance the performance. We also observed that shortest tour heuristic proposed in \cite{joshi} which selects the shortest tour among the set of B complete tours also improve the output performance of the model. 

\rev{Traditional optimization-based solvers, such as Concorde \cite{concorde}, still outperform neural network models in terms of performance. However, it is noteworthy that neural network models exhibit faster inference times than Concorde.} %revision
\rev{For example, Concorde takes approximately 16 seconds to solve 100 TSP instances of TSP100 whereas the proposed model takes approximately 11 seconds.} %revision

Our results show that the proposed model is successful in significantly reducing memory consumption and inference time. Our model is also based on a standard Transformer model with multiple layers of transformer blocks and adds a CNN embedding layer to extract spatial features. We plan to apply various lightweight techniques for Transformer-based models such as proposed in \cite{star-transformer,longformer,linformer,delight,informer} and also for CNN \cite{mobilenet1,mobilenet2,shufflenet1,shufflenet2}.

\rev{A Star-Transformer model based on a star-shaped topology has been shown to be an effective lightweight techniques for many NLP tasks \cite{star-transformer}. We plan to achieve further performance enhancement by introducing relay-nodes in the encoder used in a Star-Transformer model. Recent work has shown that ProbSparse self-attention is also effective in lightweighting Transformer-based models \cite{informer}. We also plan to improve the proposed CNN-Transformer model by introducing ProbSparse self-attention to further lightweight the model. Finally, the proposed CNN embedding layer exhibits superior performance than the linear embedding layer, but requires more computational complexity. In the subsequent study, we plan to further lightweight the CNN embedding layer by utilizing lightweight techniques in MobileNet \cite{mobilenet1,mobilenet2}.} %revision

\section{Conclusion}\label{sec:conclusion}
\rev{We proposed the first CNN-Transformer model based on partial self-attention in the decoder.}  %revision
 Our model is able to extract and learn spatial features from the input data  and produces better TSP solutions compared with the standard Transformer-based models. Our results show that our model is able to better learn local compositionality compared with the standard Transformer model. We also observed that the proposed model significantly reduces the GPU memory usage and inference time by applying  partial self-attention in the decoder. 
\rev{Our results also indicate that the proposed CNN embedding layer and partial self-attention are effective in improving performance for both random and real-world datasets.}   %revision
\rev{We found that fully-connected self-attention in the decoder can degrade performance, and performing partial self-attention only on recently visited nodes in the decoder can yield better performance than fully-connected attention models. The proposed model outperforms existing SOTA  Transformer-based models in various aspects and shows the best performance for real-world datasets.} %revision

\backmatter
\bmhead{Acknowledgments}
This work was supported in part by the National Research Foundation of Korea (NRF) Grant funded by the Korean Government (MSIT) under Grant NRF-2022R1A4A5034121. This work was also supported in part by the MSIT (Ministry of Science and ICT), Korea, under the ICAN (ICT Challenge and Advanced Network of HRD) support program (IITP-2024-RS-2023-00260175) supervised by the IITP (Institute for Information \& Communications Technology Planning \& Evaluation).

\section*{Declarations}
\bmhead{Conflict of Interests}
The authors declare that they have no conflicts of interest on this work.

\bmhead{Ethical and Informed Consent for Data Used}
Not applicable.

\bmhead{Authors Contributions}
\textbf{Minseop Jung:} Software, Methodology, Investigation, \textbf{Jaeseung Lee:} Software, Validation, \textbf{Jibum Kim:} Writing - Original Draft, Writing - Reviewing and Editing, Supervision.

\bmhead{Data Availability}
The datasets and code are available at \url{https://github.com/cm8908/CNN_Transformer3}.

%%%%%%%%%%%%%%%%%%%%%%%%%%%%%%%%%%%%%%%%%%%%%
%%%%%%%%%%%%%%%%%%%%%%%%%%%%%%%%%%%%%%%%%%%%%
%%%%%%%%%%%%%%%%%%%%%%%%%%%%%%%%%%%%%%%%%%%%%
%%%%%%%%%%%%%%%%%%%%%%%%%%%%%%%%%%%%%%%%%%%%%


%\bibliography{main-ref}


\end{document}